\documentclass[a4paper,conference]{IEEEtran}
\usepackage{ifpdf}

\usepackage{cite}

\ifCLASSINFOpdf
   \usepackage[pdftex]{graphicx}
   \graphicspath{{../pdf/}{../jpeg/}}
   \DeclareGraphicsExtensions{.pdf,.jpeg,.png}
\else
  \usepackage[dvips]{graphicx}
   \graphicspath{{../eps/}}
   \DeclareGraphicsExtensions{.eps}
\fi
\usepackage{amsmath}
\usepackage{algorithmic}

\usepackage{array}

\ifCLASSOPTIONcompsoc
 \usepackage[caption=false,font=normalsize,labelfont=sf,textfont=sf]{subfig}
\else
 \usepackage[caption=false,font=footnotesize]{subfig}
\fi

\usepackage{stfloats}
\usepackage{url}

\begin{document}
%
\title{Learning to Drive Using Sparse Imitation Reinforcement Learning}


\author{\IEEEauthorblockN{Yuci Han, Alper Yilmaz}
\IEEEauthorblockA{Photogrammetric Computer Vision Lab.\\
The Ohio State University\\
Columbus, OH 43210\\
Email: \{han.1489, yilmaz.15\}@osu.edu}
}


%


\maketitle

\begin{abstract}

In this paper, we propose Sparse Imitation Reinforcement Learning (SIRL), a hybrid end-to-end control policy that combines the sparse expert driving knowledge with reinforcement learning (RL) policy for autonomous driving (AD) task in CARLA simulation environment. The sparse expert is designed based on hand-crafted rules which is suboptimal but provides a risk-averse strategy by enforcing experience for critical scenarios such as pedestrian and vehicle avoidance, and traffic light detection. As it has been demonstrated, training a RL agent from scratch is data-inefficient and time consuming particularly for the urban driving task, due to the complexity of situations stemming from the vast size of state space. Our SIRL strategy provides a solution to solve these problems by fusing the output distribution of the sparse expert policy and the RL policy to generate a composite driving policy. With the guidance of the sparse expert during the early training stage, SIRL strategy accelerates the training process and keeps the RL exploration from causing a catastrophe outcome, and ensures safe exploration. To some extent, the SIRL agent is imitating the driving expert's behavior. At the same time, it continuously gains knowledge during training therefore it keeps making improvement beyond the sparse expert, and can surpass both the sparse expert and a traditional RL agent. We experimentally validate the efficacy of proposed SIRL approach in a complex urban scenario within the CARLA simulator. Besides, we compare the SIRL agent's performance for risk-averse exploration and high learning efficiency with the traditional RL approach. We additionally demonstrate the SIRL agent's generalization ability to transfer the driving skill to unseen environment. The supplementary material is available at \url{https://superhan2611.github.io/}.

\end{abstract}


%
\IEEEpeerreviewmaketitle

\section{Introduction}

Autonomous driving (AD) in urban environment is a challenging task, despite a lot of published research has focused on end-to-end imitation learning (IL) for the AD task \cite{codevilla2019exploring} \cite{Prakash2021CVPR} \cite{Codevilla2018} \cite{9157137} \cite{Prakash_2020_CVPR} \cite{DBLP:journals/corr/abs-1011-0686}, which aims to mimic an human expert driver's behavior. IL, however, strongly relies on the expert demonstrations that is a set of hand-crafted rules which generally performs suboptimal. Moreover, IL approach needs massive amount of human driving data which may result in poor ability to deal with out-of-distribution states and does not generalize due to driving dataset bias. Therefore, the policy learned by the IL agent usually can not surpass the expert's performance and has many limitations. 

In contrast to IL, deep reinforcement learning (DRL) allows the agent to learn on its own by trial and error and interactions with the environment. The agent learns what actions need to be taken by maximizing accumulated rewards \cite{9157239}. Hence, it will continuously learn new knowledge to improve its policy. However, the major drawback of using RL approach in autonomous driving scenario is that it needs a significantly large amounts of time to learn the driving policy from scratch, especially for the AD task with high dimensional state space along with continuous action variables. The agent will face changing traffic conditions and need to deal with lots of adverse scenarios particularly regarding the handshaking negotiations at traffic intersections, handling traffic lights and traffic signs and coping with pedestrians, cyclists, and other traffic elements. To simplify the problem, most researchers working on DRL for AD task formulate the continuous action variables into discrete variables  \cite{9157239} \cite{chen2021learning} which results in driving discomfort and unrealistic scenario handling.
Therefore, the exploration process of the RL framework is sample inefficiency that results in longer training time. The ego-vehicle needs to do large amount of dangerous exploration that may cause physical damage in a real world application.

\begin{figure}[!t]
\centering
\includegraphics[width=3.3in]{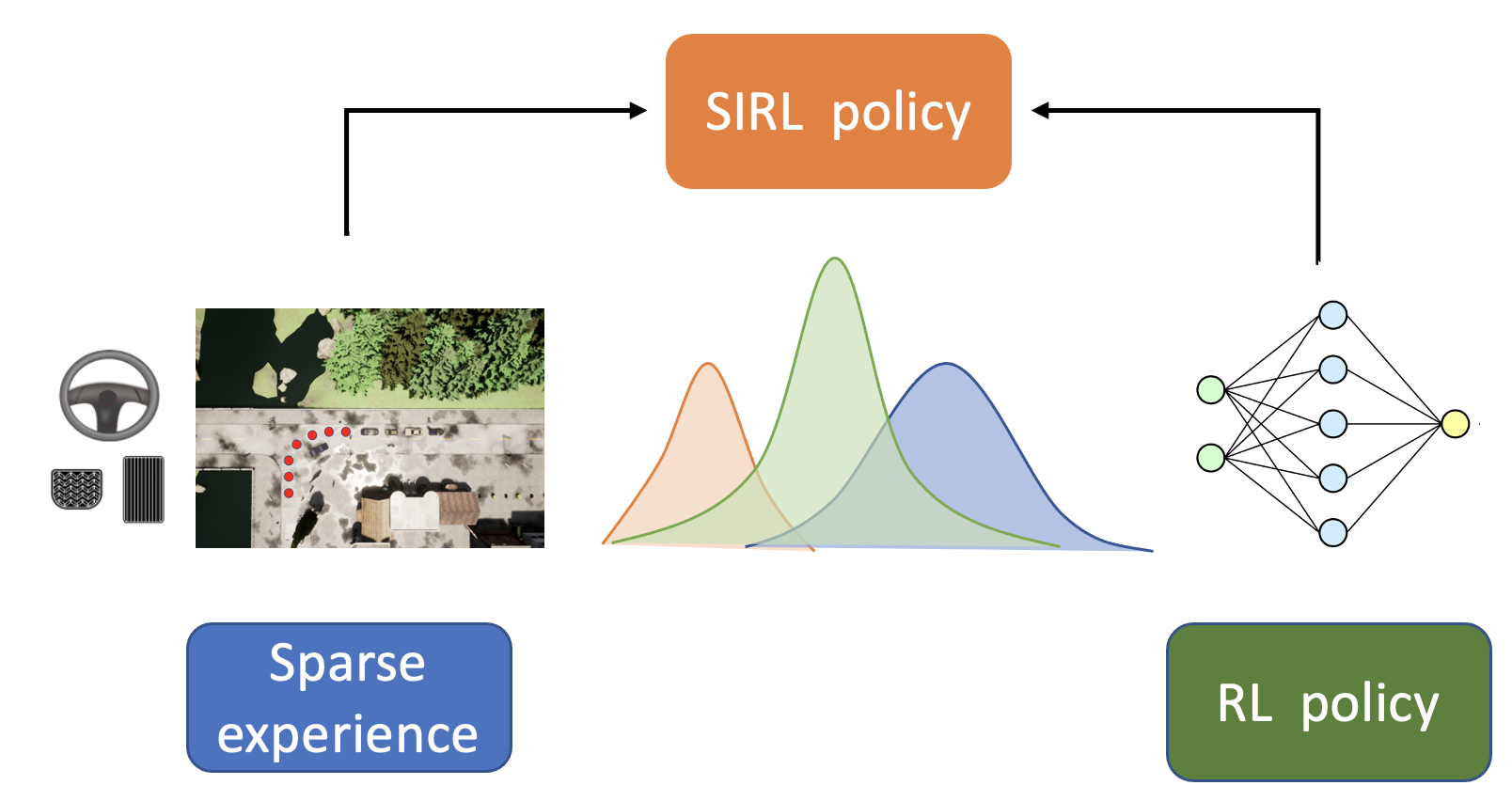}
\caption{Sparse Imitation Reinforcement Learning (SIRL): A composite policy for autonomous driving task that combines the sparse expert policy and the RL policy by fusing the output distribution of each system which ensures the efficient and safe exploration.}
\label{fig1}
\end{figure}

In this work, instead of learning the driving policy from scratch, we make use of existing hand-crafted sparse driving knowledge and introduce SIRL strategy to address these drawbacks by adapting the Bayesian Controller Fusion (BCF) approach proposed in \cite{rana2021bayesian}. The proposed approach is inspired by dual-process decision making system in human brain: habitual and deliberative. The deliberative system considers a well planned suitable action which corresponds to the RL policy in our approach whereas the habitual system responds directly to stimuli that corresponds to the intuitive sparse experience modeled from the expert. The sparse experience can give instinct reaction with it's suboptimal driving knowledge while the trained RL agent will provide a more cogitative action decision as compensation. Figure \ref{fig1} illustrates the SIRL strategy. Like human decision making mechanism, SIRL leverages the strength of each system to generate a hybrid control strategy to successfully complete the driving task \cite{Daw2005UncertaintybasedCB} \cite{Dayan2008DecisionTR}.
The sparse expert performs as control prior to aid the RL exploration process by letting the agent query an action from sparse expert at the certain state to ensure swift learning while maximizing safety. The SIRL policy combined with sparse experience has the potential to increase performance by better dealing with risk-aversity for safe behaviors in out-of-distribution states. The main contributions of our paper are as below:

\begin{itemize}
  \item We propose a novel strategy that reduces training time and ensures safe exploration in unknown environments during the reinforcement learning process for the autonomous driving task.
  \item We evaluate our ego-vehicle agent in the CARLA urban environment involving adversarial scenarios and dynamic weather conditions with continuous action space, and achieve outstanding performance. Our result shows that the driving performance of the SIRL policy outperforms the sparse expert and the RL policy individually. We improve the driving performance by 15.79\% compared with sparse expert.
  \item We make comparison with other RL approaches and showcase the advantages of our method. 
\end{itemize}

\section{related work}

In this section, we summarize the autonomous driving studies published on CARLA using imitation learning (IL) and reinforcement learning (RL) approaches. 
 
{\bf Imitation Learning (IL):}
For the autonomous driving task, IL uses the expert as the reference data to guide the learning process that maps the observation to low level control policy. IL can be trained off-line in an end-to-end manner with collected expert data. In \cite{chen2019lbc}, the authors introduced the idea of imitating a privileged agent which is trained by accessing the ground-truth data related to the environment to map camera images to way-points and achieved a performance comparable to other pipelines. Following that work, the authors of \cite{Prakash2021CVPR} utilized the idea of vision transformer to integrate RGB image and LiDAR representations using attention mechanism as the state representation for complex driving scenarios. More recently, the authors of \cite{chitta2021neat} applied an architecture combining NEAT feature representation with an implicit decoder for trajectory planning to improve IL-based autonomous driving.

{\bf Reinforcement Learning (RL):} With the introduction of deep reinforcement learning (DRL), it has become a widely used approach in robotics for navigating an autonomous ground vehicle in real world scenarios \cite{isprs-annals-V-1-2021-145-2021}. The first RL agent applied in CARLA for the AD task is based on A3C algorithm \cite{mnih2016asynchronous}. \cite{9157239} addressed the data inefficiency problem of RL by using multiple sensors to generate a fused state representation model known as implicit affordance to aid the RL learning. The authors of \cite{Prakash_2020_CVPR} made modifications on DAgger \cite{DBLP:journals/corr/abs-1011-0686} and conducted on-policy data aggregation and sampling techniques that enables the policy to generalize to new environments. The most recent research \cite{zhang2021endtoend} combined RL and IL approaches together. Instead of imitating the CARLA expert, they first trained an RL expert by accessing the ground truth of the environment and let the agent imitate the RL coach's behavior which provides stronger advice for IL process. \cite{inbook} developed a Controllable Imitative Reinforcement Learning model (CIRL). which is also based on imitative reinforcement learning. CIRL explores constrained action space guided by encoded experiences that imitate human demonstrations, using Deep Deterministic Policy Gradient (DDPG).

\begin{figure*}[!htbp]
\centering
\subfloat[CARLA route]{\includegraphics[width=1.7in]{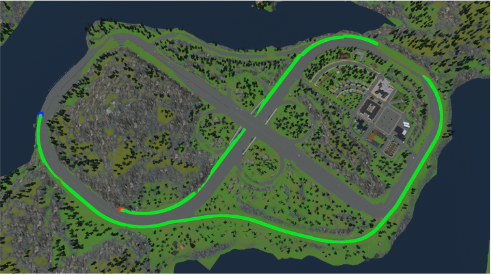}%
\label{fig_first_case}}
\hfil
\subfloat[Camera observation]{\includegraphics[width=5.1in]{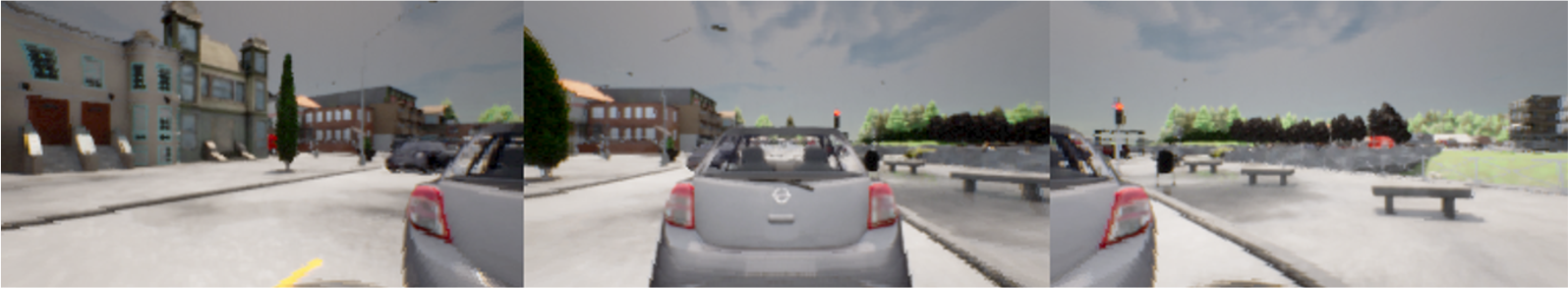}%
\label{fig_second_case}}
\caption{(a) Illustration of CARLA predefined route. (b) The ego-vehicle has forward facing RGB camera placed at the middle, left and right.}
\label{fig2}
\end{figure*}

\section{method}

This section introduces the proposed SIRL approach by setting up the problem formulation and discussing the reinforcement learning setup. 

\subsection{Problem Formulation}

Proposed SIRL algorithm is a control strategy for CARLA autonomous driving task that adopts the Bayesian controller fusion (BCF) algorithm proposed in \cite{rana2021bayesian}. In the SIRL architecture, we introduce the existing hand-crafted sparse driving experience $\psi$ as control prior to provide guidance to the RL agent. Compared with dense human expert used in former studies, the control prior is a sparse and suboptimal policy. The other component of SIRL method is the RL policy $\pi$. During the training process, it learns not only the driving ability, but also a compensation policy to the control prior. By combining the control prior $\psi(a|s)$ with a learned policy $\pi(a|s)$, SIRL method generates a composite policy $\phi(a|s)$ during both training and deployment. In contrast to train a RL agent learning the driving policy from scratch, we focus on developing the strategy to embed an existing control prior to the driving policy and let the agent imitate this expert to achieve the following behaviors:

\begin{itemize}
  \item Allowing the sparse experience to guide the early learning process thereby significantly improving the sample efficiency and learning speed compared with traditional RL algorithm.
  \item Allowing the learned policy $\pi$ to gradually dominate the control as it gains more knowledge such that the hybrid SIRL control strategy $\phi$ eventually outperforms both the expert and RL policy when used independently. 
  \item Ensuring safe exploration and deployment while interacting with the environment. The SIRL policy biases towards the control prior during the early training stage which is suboptimal but risk-averse hence reduces the learning cost.
\end{itemize}

\subsection{Reinforcement Learning Setup}

{\bf Environemnt.} We conduct our experiment in the CARLA \cite{dosovitskiy2017carla} environment. The objective of our ego-vehicle is to drive through a set of predefined routes (see Figure. \ref{fig2}(a)). For each route, agents will be initialized at a starting point and directed to drive to a destination point. Routes will be generated in a variety of areas of the road network, including freeways, urban scenes, and residential districts. Agents are evaluated in a variety of weather conditions, including daylight scenes, sunset, rain, fog, and night and multiple traffic situations.

{\bf Observation.} We use forward facing RGB camera placed at the middle, left and right of the ego-vehicle (see Fig. \ref{fig2}(b)). The raw image resolution is 768x144. We use ImageNet pretrained ResNet-50 encoder \cite{he2015deep} to produce 2048 dimensional feature as the input state to the RL training process.

{\bf Action and Reward shaping.} The actions of the ego-vehicle are steering $\in [-1, 1]$, throttle $\in [0, 1]$ and brake $\in [0, 1]$. The outputs of the SIRL policy are steering and throttle values. We set a threshold for brake, for example if the throttle value is less than 0.005, the brake value is True and False otherwise. The training reward is calculated using the waypoint path planning API in CARLA system. It provides the desired location and desired shape (vehicle rotation) for every timestamp for a given route and scenario. We compare the current position and the shape of our ego-vehicle with desired position and shape, if the difference is larger than the threshold, the reward will be -0.01 respectively and 0 otherwise. For 100\% route completion, we add 20 points to the total reward.

\subsection{Sparse Imitation Reinforcement Learning}

SIRL combines policies from two control strategies: an RL policy $\pi(a|s)$ and a sparse expert policy $\psi(a|s)$. For AD task, we conjecture that both $\pi(a|s)$ and $\psi(a|s)$ follow Gaussian distribution. These two distributions give two independent action estimations which provide the steering and throttle controls. The derivation follows \cite{rana2021bayesian}, and we fuse the RL policy and control prior using normalized product. The hybrid policy $\phi(a|s)$ is also a Gaussian distribution over two actions. The hybrid estimates of an action $a$ is:

\begin{equation}
p(a|\theta _{\pi }, \theta _{\psi })=\frac{p(\theta _{\pi }, \theta _{\psi }|a)p(a)}{p(\theta _{\pi }, \theta _{\psi })}.
\label{1} 
\end{equation} 
We have the RL policy $\pi(a|s)\approx p(a|s, \theta _{\pi})$ and the control prior $\psi(a|s)\approx p(a|s, \theta _{\psi})$, where $\theta _{\pi}$ and $\theta _{\psi}$ denote the parameters for each driving policy distribution. Assuming the $p(a|s, \theta _{\pi})$ and $p(a|s, \theta _{\psi})$ are statistical independent, the likelihood estimation of $p(\theta _{\pi }, \theta _{\psi }|a)$ becomes:

\begin{equation}
\begin{aligned} 
p(\theta_{\pi},\theta_{\psi}|a) &= p(\theta_{\pi}|a)p(\theta _{\psi}|a) \\
                                &= \frac{p(a|\theta_{\pi })p(\theta_{\pi})}{p(a)}\frac{p(a|\theta_{\psi})p(\theta_{\psi})}{p(a)}.
\label{2} 
\end{aligned}
\end{equation} 
We simplify the composite policy  $\phi(a|s)$ by substituting (2) into (1) and obtain:

\begin{equation}
\phi(a|s)=p(a|\theta_{\pi},\theta_{\psi}) = \eta p(a|\theta_{\pi})p(a|\theta_{\psi}),
\label{3} 
\end{equation} 

\begin{equation}
\eta = \frac{p(\theta_{\pi})p(\theta_{\psi})}{p(\theta_{\pi}, \theta_{\psi})p(a)},
\label{4} 
\end{equation} 
which is a normalized product of the RL policy and control prior. This cc. The $\mu _{\phi }$ and $\sigma _{\phi}^{2}$ are calculated based on the mean value $\mu$ and the variance value $\sigma^{2}$ of two control systems as follows:

\begin{equation}
\mu _{\phi} = \frac{\mu _{\pi}\sigma _{\psi}^{2}+\mu _{\psi}\sigma _{\pi}^{2}}{\sigma _{\psi}^{2}+\sigma _{\pi}^{2}},
\label{5} 
\end{equation}

\begin{equation}
\sigma _{\phi}^{2} = \frac{\sigma _{\psi}^{2}\sigma _{\pi}^{2}}{\sigma _{\psi}^{2}+\sigma _{\pi}^{2}}.
\label{6} 
\end{equation} 

We describe two components, sparse experience control prior and the reinforcement learning policy of the SIRL approach in the following discussion.

{\bf Sparse Experience Control Prior $\mathbf{\psi}$.}
The sparse experience is the existing suboptimal hand-crafted driving policy which we generate from the CARLA Autopilot API. It provides expert control commands at certain locations for example every 5 meters in our case since we expect this expert to provide as much experience as possible. In CARLA simulator, the ego-vehicle is sensitive to the control turbulence, therefore, to successfully complete the driving task, the agent needs continuous optimal policy at every time step. Even though the sparse experience control prior is an ineffective policy since it is sparse and suboptimal, it provides guidance at intervals of every 5 to 10 meters to let the SIRL agent imitate and make improvement based on the control prior. 

The control prior follows the Gaussian distribution $\mathcal{N}(\mu_{\psi },\sigma_{\psi}^{2})$. The mean value $\mu_{\psi}$ is calculated by propagating noise from the sensor measurements to the output actions using Monte Carlo sampling. $\sigma _{\psi}^{2}$ is defined as a hyper-parameter to control the impact of the sparse expert has on guiding the exploration process. Small $\sigma_{\psi}^{2}$ value will let the hybrid distribution bias to the control prior, hence, it increases the impact of prior knowledge and constrains the exploration variety. As a result, the SIRL agent will always be prone to the chosen action similar to the control prior. The region of exploration decrease with smaller $\sigma _{\psi}^{2}$ and vice versa. We explore the impact of the control prior in the experiment section.

\begin{figure}[!t]
\centering
\includegraphics[width=3.5in]{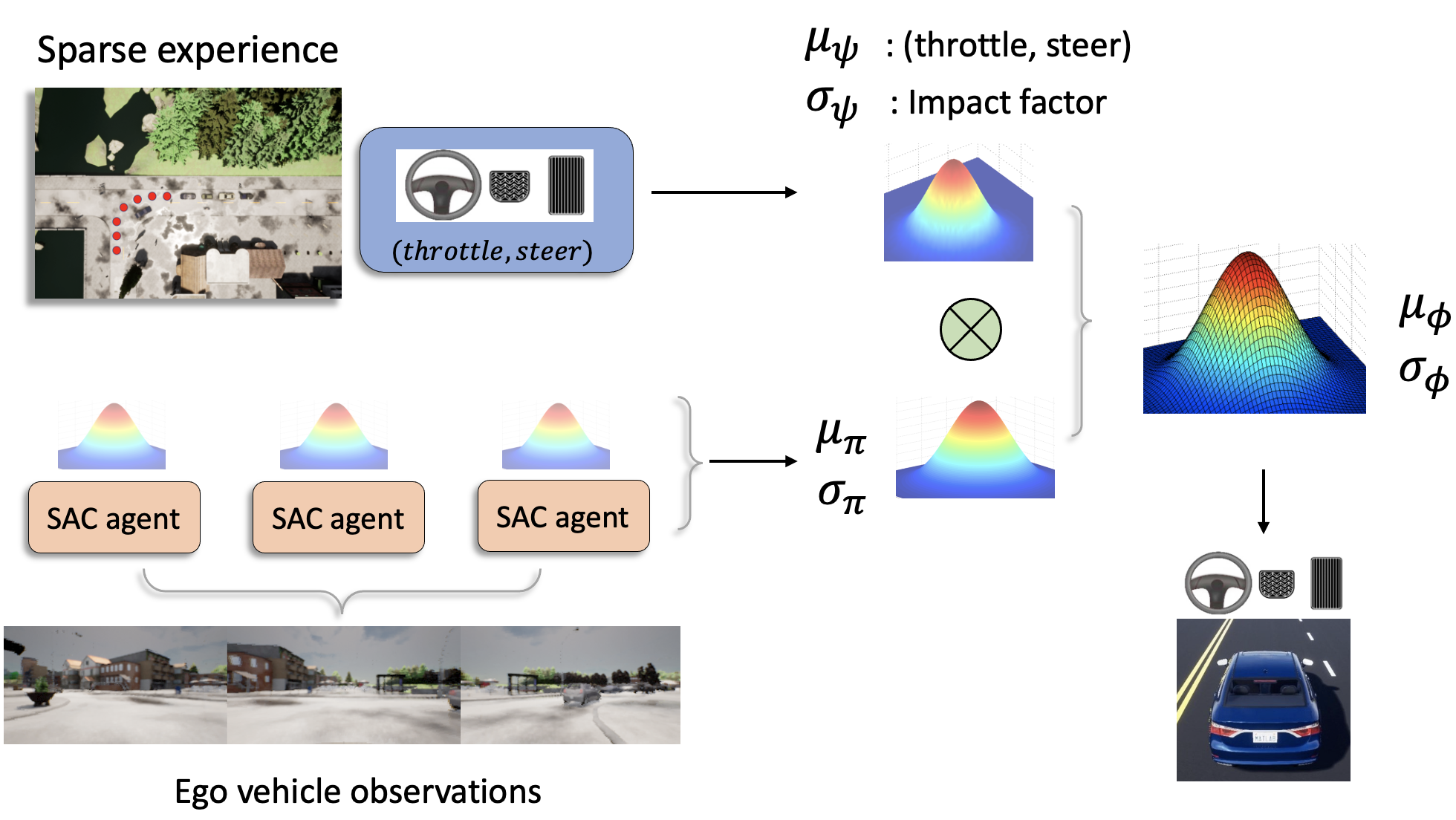}
\caption{Schematic illustration of the SIRL model. The sparse expert follows the Gaussian distribution $\mathcal{N}(\mu_{\psi },\sigma_{\psi}^{2})$.  The RL policy combines the outputs of 3 SAC agents into a uniformly weighted Gaussian mixture model $\mathcal{N}(\mu_{\pi }, \sigma_{\pi}^{2})$. SIRL method combines policies from these two control strategies and will also be a Gaussian distribution $\phi (a|s) \sim \mathcal{N}(\mu _{\phi }, \sigma _{\phi}^{2})$.}
\label{fig4}
\end{figure}

{\bf Reinforcement Learning Policy $\mathbf{\pi}$.}
The SIRL method implements a Deep Reinforcement Learning backbone. We formulate the driving task as a Markov Decision Process (MDP) with state space $\mathcal S$ and action space $\mathcal A$ and implement the RL framework. During the exploration, we generate a trajectory which is a sequence of states, actions and rewards $\mathcal T$ = \{ $ s_t, a_t, r_t,...,a_{T-11}, r_{T-1},  s_{T-1} $ \}. The RL agent learns an optimal policy that maximize the cumulative reward. We assume the normally distributed $\mathcal{N}(\mu_{\pi }, \sigma_{\pi}^{2})$ output from this RL policy with mean $\mu_{\pi }$ and variance $\sigma_{\pi}^{2}$. The method we use to train the RL agent is the Soft Actor Critic (SAC) algorithm \cite{haarnoja2018soft} which is a state of the art algorithm for off-policy stochastic RL. Instead of training a single agent, we train multiple agents to learn the driving policy simultaneously and integrate these in a uniformly weighted Gaussian mixture model to calculate the $\mu_{\pi }$ value and the $\sigma_{\pi}^{2}$ value and generate a single Gaussian distribution (see Figure.\ref{fig4}) as the final policy of RL control system and feed it to the SIRL policy fusion process. 

\section{Experiments}

In this section, we provide comparative analysis of the proposed SIRL and other baseline algorithms for the autonomous driving task in Carla 0.9.10 simulation environment. Our training and test are all completed in a CARLA European town with one-lane roads and T-junctions \cite{zhang2021endtoend}. We choose five different routes for training and testing generalization of the model. We conduct five types of experiments to validate our approach. First, we evaluate the driving performance of our SIRL agent and compare with other baselines. Second, we explore the accelerated-learning and risk-averse exploration ability of SIRL approach and show the advantage over other methods. Third, we analyze the impact of the sparse expert to see how it guides the learning process. Finally, we conduct the driving task in unseen route to test the generalization ability of our SIRL agent. 

\subsection{Driving Task and Training Settings}

We consider the AD task as navigating along the predefined route with low level control commands including steering, throttle and brake. The route is defined by a sequence of GPS coordinates provided by the CARLA simulator. Our simulation consists the following scenarios:
\begin{itemize}
    \item obstacle avoidance,
    \item vehicle running red lights, 
    \item pedestrian emerging to cross the road from occluded regions at any arbitrary locations, 
    \item following dynamic traffic flows and 
    \item cope with high density traffic and dynamic agents.
\end{itemize}

The parameters of all the networks are randomly initialized. The input of the SAC algorithm is the ResNet-50 extracted features from 768x144 raw image. All experiments are trained on a NVIDIA 3070 GPU with Adam optimizer. While training the networks, we use learning rate $\eta = 10^{-3}$ and the discount factor $\gamma = 0.99$. For RL system, we train 3 SAC agents simultaneously. The episode terminates once the ego-vehicle deviates from desired position and rotation more than 5 meters in total. 

\subsection{Baselines}

The RL baselines of our experiment are based on \cite{rana2021bayesian} and include the following approaches:
\begin{itemize}
    \item {\bf Sparse Expert:} Our sparse expert control prior $\psi$ described in section III.
    \item {\bf Dense Expert:} The Autopilot API in the CARLA environment which has expert driving performance using the route planner module.
    \item {\bf Residual RL:} Residual Reinforcement learning algorithm proposed by \cite{johannink2018residual}. The policy of residual RL combines the conventional control methods and the residual which is solved with RL.
    \item {\bf KL Regularised RL:} Modified SAC algorithm proposed by \cite{pertsch2020accelerating}. The objective of this method is the KL divergence between the distribution of the learned policy and the control prior. The agent is trained aims at imitating the expert's behavior. 
    \item {\bf SAC:} SAC algorithm optimizes a stochastic policy in an off-policy way by adding an entropy regularization.
\end{itemize}

\begin{figure*}[!t]
\centering
\includegraphics[width=7in]{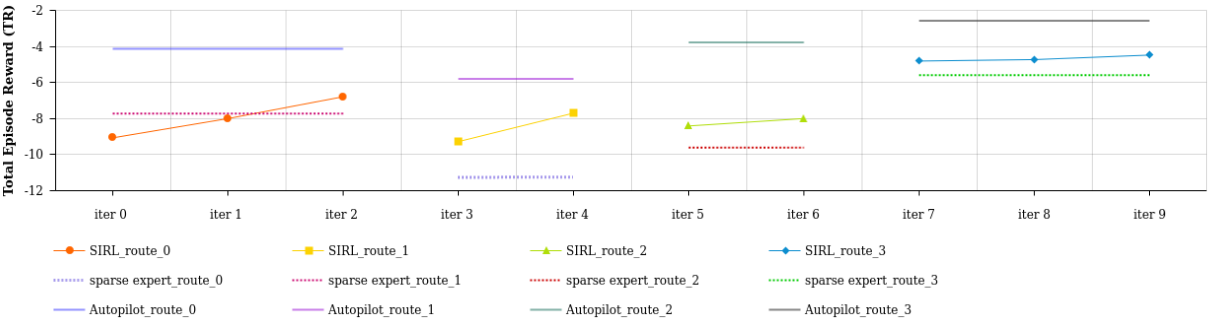}
\caption{Training process tracking the agent's episode reward for each training iteration}
\label{fig5}
\end{figure*}

\begin{table*}
\parbox{.65\linewidth}{
\centering
\caption{Driving Performance}
\label{table_1}
\begin{tabular}{c|cc|cc|cc|cc}
\hline
\textbf{Method}      & \multicolumn{2}{c|}{\textbf{Route00}} & \multicolumn{2}{c|}{\textbf{Route01}} & \multicolumn{2}{c|}{\textbf{Route02}} & \multicolumn{2}{c}{\textbf{Route03}}        \\ \hline
                     & TR                & RC                & TR                & RC                & TR                & RC                & TR             & RC                         \\ \hline
\textbf{SIRL (ours)} & \textbf{12.30}    & \textbf{100\%}    & \textbf{15.52}    & \textbf{100\%}    & \textbf{12.00}    & \textbf{100\%}    & \textbf{13.21} & \textbf{100\%}             \\
Sparse Expert        & 8.73              & 100\%             & 14.41             & 100\%             & 10.39             & 100\%             & 12.27          & 100\%                      \\
SAC                  & -                 & 5.42\%            & -                 & 8.15\%            & -                 & 8.60\%            & -              & \multicolumn{1}{l}{4.80\%} \\
Residual RL          & -                 & 5.55\%            & -                 & 7.94\%            & -                 & 10.01\%           & -              & \multicolumn{1}{l}{12.3\%} \\
KL Regularised       & -                 & 4.46\%            & -                 & 5.81\%            & -                 & 6.25\%            & -              & \multicolumn{1}{l}{6.48\%} \\ \hline
Expert               & \textit{14.2}     & \textit{100\%}    & \textit{17.43}    & \textit{100\%}    & \textit{16.23}    & \textit{100\%}    & \textit{15.87} & \textit{100\%}             \\ \hline
\end{tabular}
}
\hfill
\parbox{.35\linewidth}{
\centering
\caption{Impact of control prior}
\label{table_2}
\begin{tabular}{c|cl|cl}
\hline
\textbf{\begin{tabular}[c]{@{}c@{}}Sigma Value\\ (Control Prior)\end{tabular}} & \multicolumn{2}{c|}{\textbf{RC}} & \multicolumn{2}{c}{\textbf{TR}} \\ \hline
0.1                                                                            & \multicolumn{2}{c|}{100\%}       & \multicolumn{2}{c}{13.63}       \\
0.05                                                                           & \multicolumn{2}{c|}{100\%}       & \multicolumn{2}{c}{14.89}            \\
\textbf {0.01}                                                                           & \multicolumn{2}{c|}{100\%}       & \multicolumn{2}{c}{\textbf{15.19}}       \\
0.005                                                                          & \multicolumn{2}{c|}{100\%}       & \multicolumn{2}{c}{14.87}       \\
0.001                                                                          & \multicolumn{2}{c|}{100\%}       & \multicolumn{2}{c}{14.37}       \\ \hline
\end{tabular}
}
\end{table*}

\subsection{Driving Performance}

In our first experiment, we evaluate the driving performance of the SIRL policy and compare it with other baselines. Throughout the experiment, we use two criteria for evaluation: (1) {\bf Total Reward (TR)}, which is the accumulated reward for the episode that accounts for deviation from the desired position at each time step, (2) {\bf Route Completion Rate (RC)}, percentage of completed route distance. This metric is introduced by CARLA LeaderBoard. Figure \ref{fig5} illustrates the training process of the SIRL agent for four different routes. For each training iteration, we collect driving experience \{ $ s_t, a_t, r_t, s_{t+1}, a^{prior}_t, a^{prior}_{t+1}$ \} for the whole episode and store it in the replay buffer. While updating the RL policy, we sample batch from the buffer and train the agent in an off-policy manner. From Figure. \ref{fig5}, we observe that the SIRL agent achieves similar performance to the sparse expert at the second training iteration and outperforms it in the following iterations. 

Table \ref{table_1} shows the comparison of the driving performance between different baselines including the spares expert and dense expert (CARLA Autopilot). For comparison, we train the RL baselines and SIRL agent for the same training steps. As shown in Table \ref{table_1}, our SIRL agent has the best performance that achieves a higher total reward and route completion rate than other baselines especially compared with other RL approaches. Even under the same training condition, other RL baselines perform poorly for all evaluation routes with low route completion values. Besides that, we show the proposed SIRL agent improve from sparse expert by 15.79\% in average in terms of TR value while other RL agents overall fail to learn the AD task. Compared with sparse expert and pure SAC agent, our SIRL agent shows the behavior that as it gains more knowledge, the RL policy $\pi$ gradually compensates the control and the composite policy $\phi$ has significant improvement than the sparse expert and pure SAC agent independently. We can speculate that this improvement in performance between the two approaches is related to the performance gap between the sparse expert and the optimal expert policy of the driving task. As for the reason of poor performance of the pure SAC agent, we consider this is due to the inefficient exploration of the RL agent, because in the training process, the agent needs to collection driving experience to build the replay buffer for off-policy learning. Traditional RL method learns from scratch hence easily goes into catastrophic situations and triggers the end of the episode which severely impedes the exploration process, hence, it needs longer time to collect experience data to achieve better performance.

\subsection{Accelerated Learning}

\begin{figure*}%
    \centering
    \subfloat[\centering SIRL policy]{{\includegraphics[width=4.4cm]{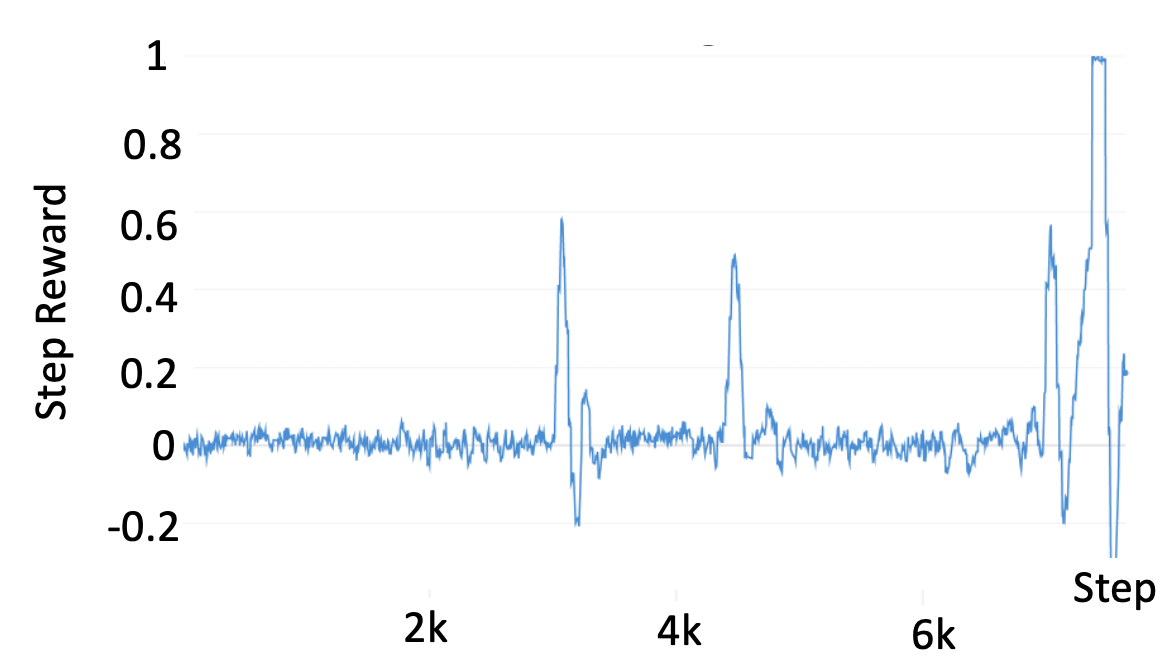} }}%
    \subfloat[\centering KL-Regularized policy]{{\includegraphics[width=4.4cm]{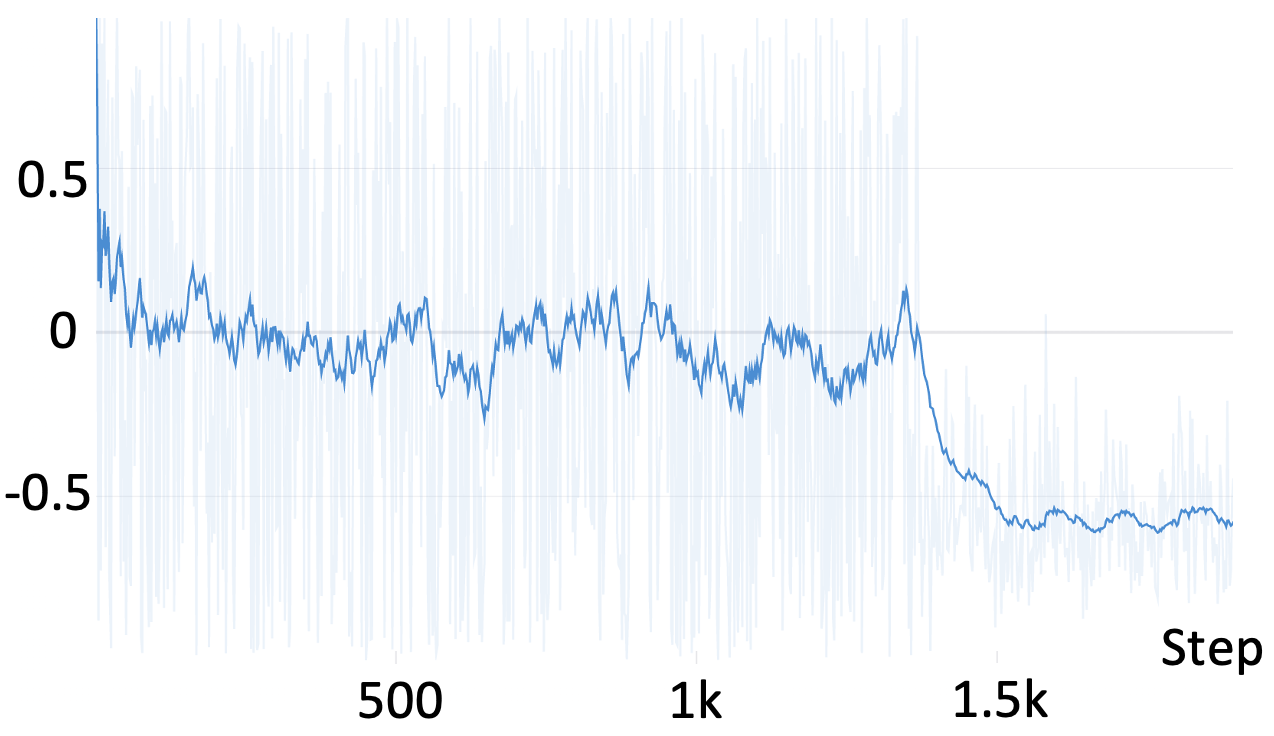} }}%
    \subfloat[\centering Residual RL policy]{{\includegraphics[width=4.4cm]{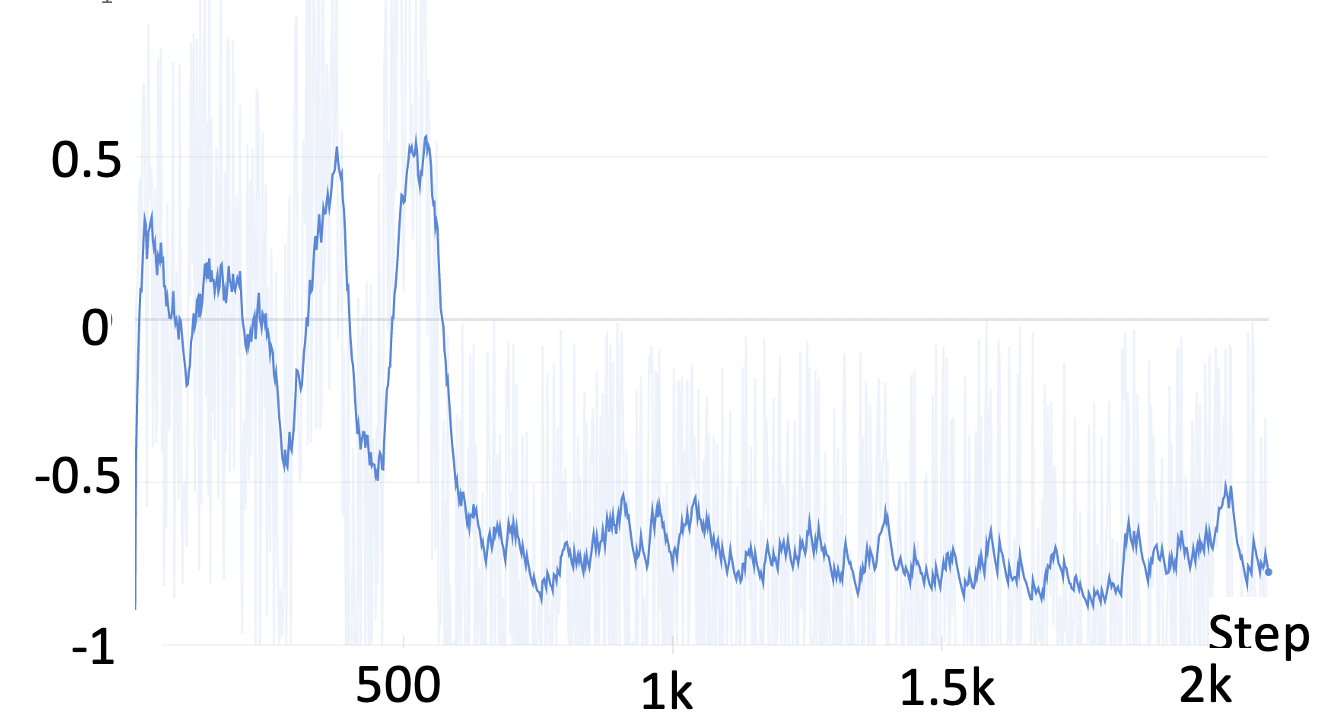} }}%
    \subfloat[\centering SAC policy]{{\includegraphics[width=4.4cm]{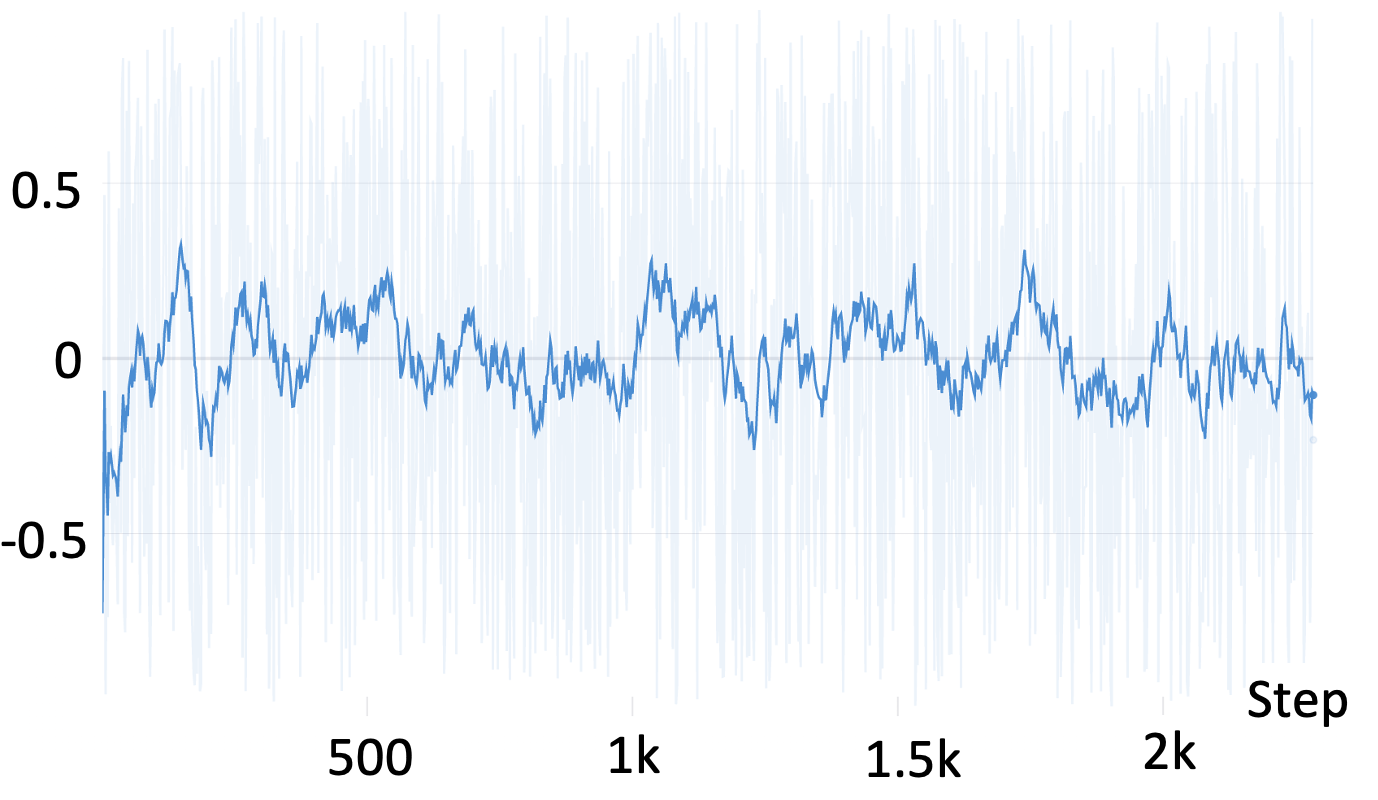} }}%
    \qquad
    \subfloat[\centering SIRL (control prior)]{{\includegraphics[width=4.4cm]{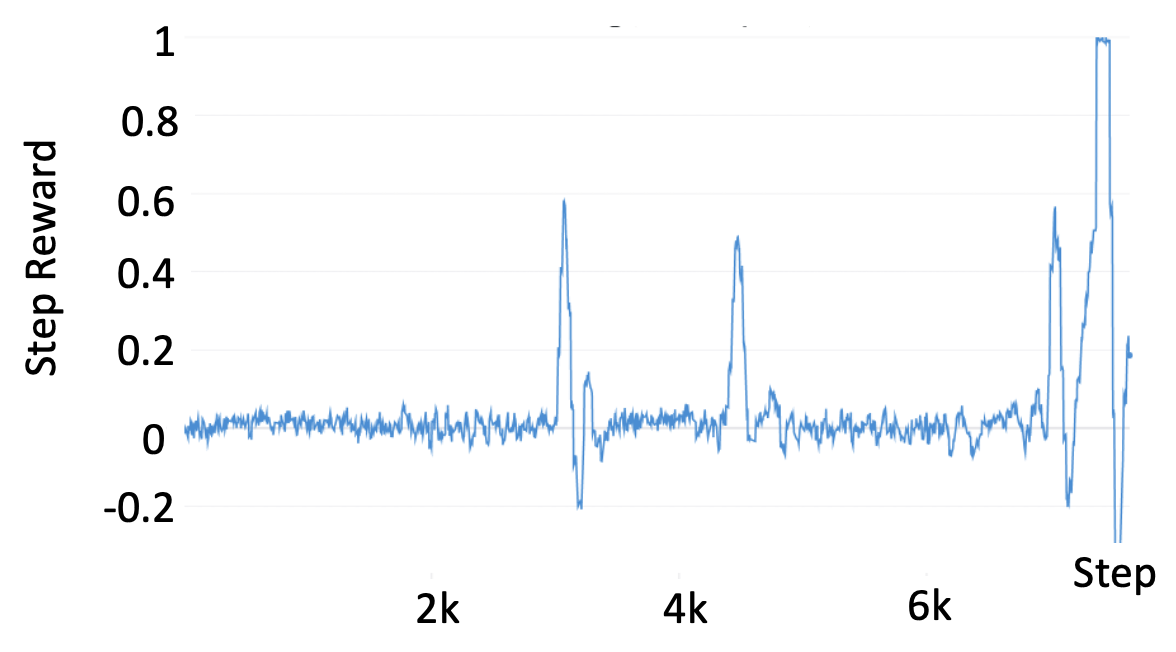} }}%
    \subfloat[\centering KL-Regularized (control prior)]{{\includegraphics[width=4.4cm]{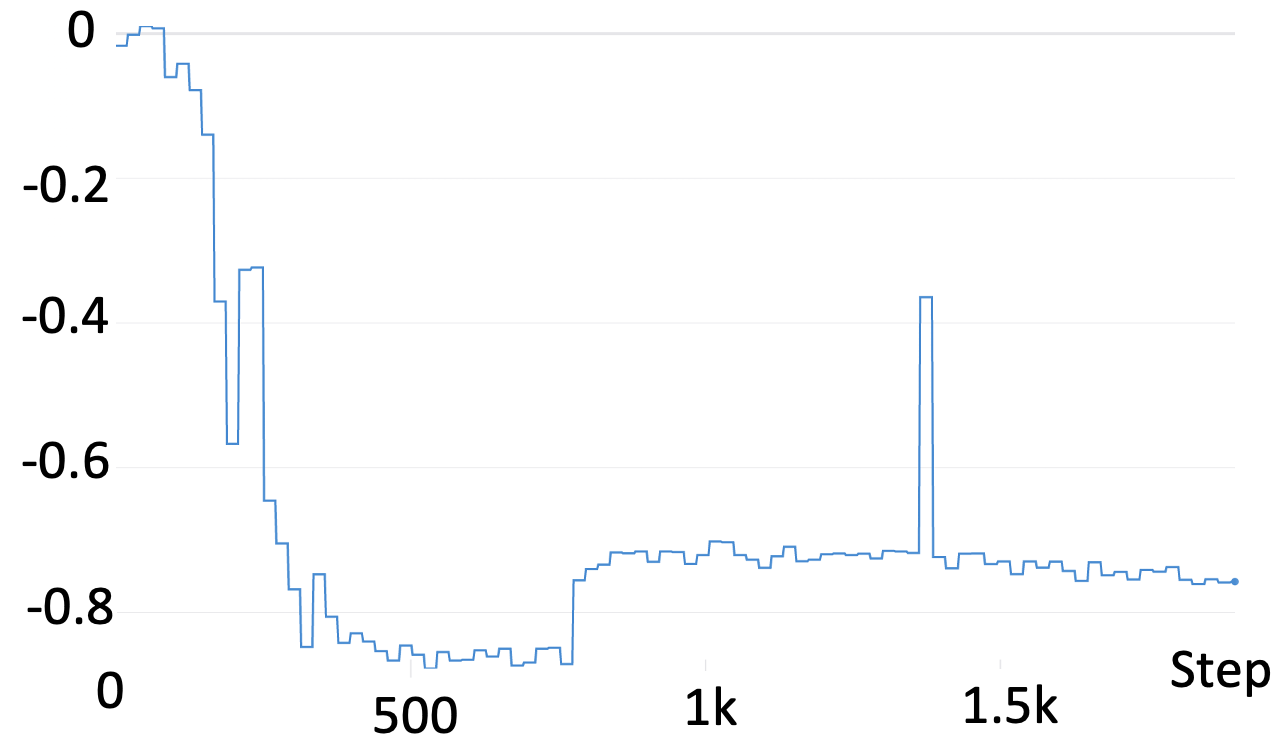} }}%
    \subfloat[\centering Residual RL (control prior)]{{\includegraphics[width=4.4cm]{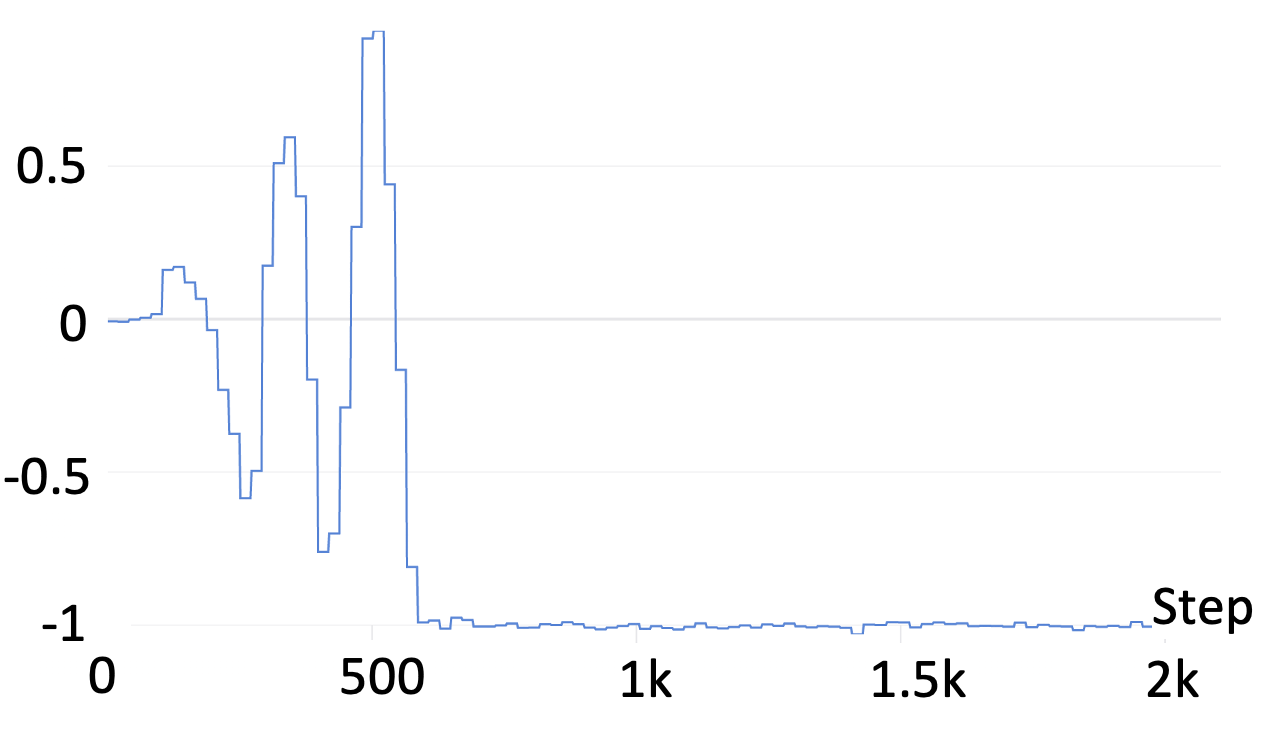} }}%
    \subfloat[\centering SAC (control prior)]{{\includegraphics[width=4.4cm]{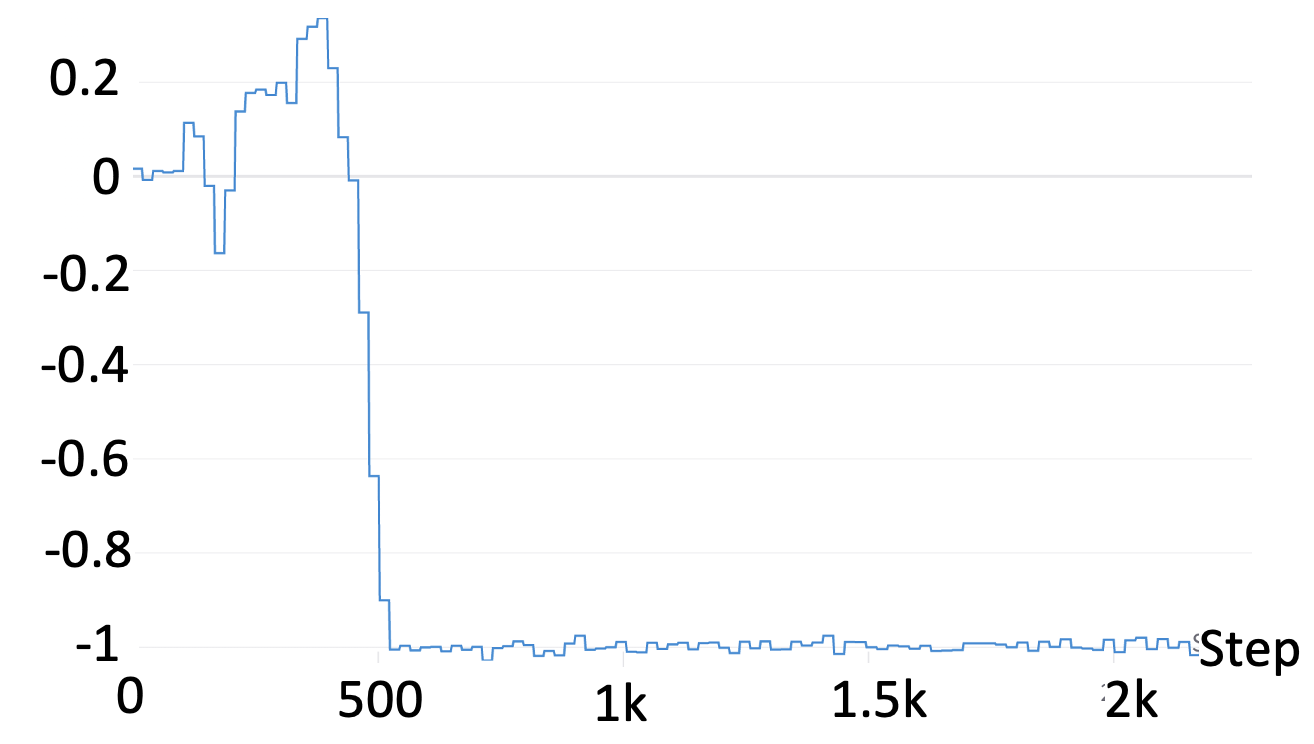} }}%
    \caption{Policy distribution of different baselines for the first training episode compared with expert prior policy accordingly. SIRL agent shows the strongest imitation performance and the accelerated learning ability towards the expert policy. KL-Regularized RL and Residual RL also show the trend to converge to an expert policy  when compared to SAC alone.}%
    \label{fig6}%
\end{figure*}

In this experiment, we demonstrate the SIRL agent's ability to accelerate the learning process. For easy comparison, we train all the baselines for one same route to observe the early learning behavior of different approaches. The results in Table \ref{table_3} demonstrate the driving performance for the first training episode. During the early stage of training, the SIRL policy biases the distribution $\phi(a|s)$ towards the sparse expert prior $\psi(a|s)$ aiming at yielding higher reward. This allows the agent to collect the driving experience efficiently and quickly gains knowledge for the RL agent to improve its driving ability. As shown in Table \ref{table_3}, With the guidance of the sparse expert, the SIRL agent is able to complete the full route in the first training episode while the Residual RL and regular SAC agent which learn from scratch achieve less than 5\% route completion rate. Figure \ref{fig6} illustrates the learned policy distribution of each approach compared with expert prior accordingly. Compared with other approaches, SIRL agent shows the strongest imitation performance and the accelerated learning ability towards the expert policy. KL-Regularized RL and Residual RL also show the trend to converge to an expert policy when compared to SAC alone. However, they all have lower learning efficiency compared with SIRL agent and overall fail the AD task in the early training stage.

\begin{table*}
\parbox{.65\linewidth}{
\centering
\caption{Driving and Infraction Performance for the first training episode (accelerated learning and risk-averse exploration) }
\label{table_3}
\begin{tabular}{c|cc|cc|cc|cc}
\hline
\textbf{}            & \multicolumn{2}{c|}{\textbf{Route Completion}} & \multicolumn{2}{c|}{\textbf{Outside Route Lanes}} & \multicolumn{2}{c|}{\textbf{Collision}} & \multicolumn{2}{c}{\textbf{Running Red Light}} \\ \hline
\textbf{SIRL (ours)} & \multicolumn{2}{c|}{\textbf{100\%}}            & \multicolumn{2}{c|}{\textbf{0\%}}                 & \multicolumn{2}{c|}{{4}}         & \multicolumn{2}{c}{\textbf{0}}                 \\
KL regularised       & \multicolumn{2}{c|}{2.42\%}                    & \multicolumn{2}{c|}{22.26\%}                      & \multicolumn{2}{c|}{\textbf{1}}                  & \multicolumn{2}{c}{\textbf{0}}                          \\
Residual RL          & \multicolumn{2}{c|}{4.94\%}                    & \multicolumn{2}{c|}{2.26\%}                       & \multicolumn{2}{c|}{\textbf{1}}                  & \multicolumn{2}{c}{\textbf{0}}                          \\
SAC                  & \multicolumn{2}{c|}{2.68\%}                    & \multicolumn{2}{c|}{32.2\%}                       & \multicolumn{2}{c|}{2}                  & \multicolumn{2}{c}{1}                          \\ \hline
\end{tabular}
}
\hfill
\parbox{.35\linewidth}{
\centering
\caption{Generalization Performance}
\label{table_4}
\begin{tabular}{c|cc}
\hline
\textbf{Method}      & \multicolumn{2}{c}{\textbf{Test Route}} \\ \hline
                     & TR                 & RC                 \\ \hline
\textbf{SIRL (ours)} & \textbf{12.90}     & \textbf{100\%}     \\
Sparse Expert        & 12.27              & 100\%              \\
SAC                  & -                  & 6.78\%             \\
Residual RL          & -                  & 7.62\%             \\
KL Regularised       & -                  & 4.96\%                  \\ \hline
Expert               & \textit{17.58}     & \textit{100\%}     \\ \hline
\end{tabular}
}
\end{table*}

\subsection{Risk-Averse Exploration}

A key advantage of SIRL approach is the risk-averse exploration ability. We investigate the safe exploration ability of our SIRL agent in the early training period and compare it with other methods. This advantage has significant benefit to reduce the exploration cost when gradually transition towards training the agent in the real world for the autonomous driving task. We evaluate the safe-exploration ability by investigating the obstacle collisions, running red light and running outside route lanes experienced by the agent during the exploration process. All baselines are trained under the same conditions. The result is shown in table \ref{table_3}, during the first training episode, the SIRL agent achieves 100\% route completion with four  collisions, zero time running red light and 0\% running outside route lanes with full route completion. We attribute this to the sparse expert's guidance to the composite policy that allows the agent to query the prior knowledge when encounter these unsafe situations. In contrast, all other baselines experience several collisions and running outside route lanes throughout the driving distance less than 5\% of the full route length which implies higher exploration risk.

\subsection{Impact of Sparse Expert}

In this section, we analyze the impact of the sparse expert prior has on guiding the training process. The impact of the sparse expert prior is reflected by its standard deviation value ($\sigma _{\psi}$). With small $\sigma _{\psi}$ value, the SIRL policy is biased towards a solution similar to the control prior distribution. As the result shown in Table \ref{table_2}, there is an optimal $\sigma _{\psi}$ value that can balance the exploration and exploitation in an optimal manner. Small $\sigma _{\psi}$ value amplifies the influence of the control prior hence impedes the exploration process which is critical for RL agent to learn and make improvement beyond the control prior. On the contrary, large $\sigma _{\psi}$ value reduce the impact of the control prior, as a result, the agent is lack of guidance during the training and gains lower reward due to random exploration.

\subsection{Generalization Ability}

We additionally investigate the generalization ability of our SIRL policy to drive in unseen route. Table \ref{table_4} provides the detailed driving performance for the test route. Our SIRL agent achieves 100\% route completion rate and performs comparably well to the sparse expert. However, other RL baselines fail to generalize the policy to the unseen route with less than 10\% route completion rate. Most notably, with the composite policy of our strategy, the SIRL policy shows an improvement of 5.13\% in average in terms of total reward compared with the sparse expert. But compared with dense expert, there are still gaps to be addressed.

\section{Conclusion}

In this work, we present SIRL, a RL based control strategy for autonomous driving task using continuous driving actions. Our approach provides an effective way to imitate the existing driving strategy by combining the existing hand-crafted suboptimal sparse driving policy with a RL model to generate a hybrid control policy. We show that our strategy not only accelerates the training process but also ensures risk-averse exploration guided by the classical controllers. Besides that, we show our agent successfully complete given autonomous driving task and the SIRL agent substantially outperforms the sparse expert by a margin of 15.79\% improvement computed in terms of total reward. We demonstrate the generalization ability of our SIRL policy to unseen environment by achieving 100\% route completion and attaining a higher total reward compared with sparse expert. Given that our approach is flexible and generic, in the future work, it would be easy to replace the sparse expert with other control prior method for AD task such as CILRS \cite{codevilla2019exploring}, CIL \cite{Codevilla2018}, CAL \cite{sauer2018conditional}, SAM \cite{zhao2020sam} providing a generic framework with a generalized for the autonomous driving ability.

\section{Acknowledgement}

The study conducted in this paper is funded by the Army Research Office award number W911NF2110356 as basic scientific research. The authors thank Charles Toth and John Anderson for advice on experimental design.






\bibliographystyle{IEEEtran}
\bibliography{IEEEexample}

\begin{thebibliography}{10}
\providecommand{\url}[1]{#1}
\csname url@samestyle\endcsname
\providecommand{\newblock}{\relax}
\providecommand{\bibinfo}[2]{#2}
\providecommand{\BIBentrySTDinterwordspacing}{\spaceskip=0pt\relax}
\providecommand{\BIBentryALTinterwordstretchfactor}{4}
\providecommand{\BIBentryALTinterwordspacing}{\spaceskip=\fontdimen2\font plus
\BIBentryALTinterwordstretchfactor\fontdimen3\font minus
  \fontdimen4\font\relax}
\providecommand{\BIBforeignlanguage}[2]{{%
\expandafter\ifx\csname l@#1\endcsname\relax
\typeout{** WARNING: IEEEtran.bst: No hyphenation pattern has been}%
\typeout{** loaded for the language `#1'. Using the pattern for}%
\typeout{** the default language instead.}%
\else
\language=\csname l@#1\endcsname
\fi
#2}}
\providecommand{\BIBdecl}{\relax}
\BIBdecl

\bibitem{codevilla2019exploring}
F.~Codevilla, E.~Santana, A.~M. L{\'o}pez, and A.~Gaidon, ``Exploring the
  limitations of behavior cloning for autonomous driving,'' \emph{International
  Conference on Computer Vision(ICCV)}, 2019.

\bibitem{Prakash2021CVPR}
A.~Prakash, K.~Chitta, and A.~Geiger, ``Multi-modal fusion transformer for
  end-to-end autonomous driving,'' in \emph{Conference on Computer Vision and
  Pattern Recognition (CVPR)}, 2021.

\bibitem{Codevilla2018}
F.~Codevilla, M.~M{\"u}ller, A.~L{\'o}pez, V.~Koltun, and A.~Dosovitskiy,
  ``End-to-end driving via conditional imitation learning,'' in
  \emph{International Conference on Robotics and Automation (ICRA)}, 2018.

\bibitem{9157137}
E.~Ohn-Bar, A.~Prakash, A.~Behl, K.~Chitta, and A.~Geiger, ``Learning
  situational driving,'' in \emph{2020 IEEE/CVF Conference on Computer Vision
  and Pattern Recognition (CVPR)}.\hskip 1em plus 0.5em minus 0.4em\relax Los
  Alamitos, CA, USA: IEEE Computer Society, jun 2020, pp. 11\,293--11\,302.

\bibitem{Prakash_2020_CVPR}
A.~Prakash, A.~Behl, E.~Ohn-Bar, K.~Chitta, and A.~Geiger, ``Exploring data
  aggregation in policy learning for vision-based urban autonomous driving,''
  in \emph{Proceedings of the IEEE/CVF Conference on Computer Vision and
  Pattern Recognition (CVPR)}, June 2020.

\bibitem{DBLP:journals/corr/abs-1011-0686}
\BIBentryALTinterwordspacing
S.~Ross, G.~J. Gordon, and J.~A. Bagnell, ``No-regret reductions for imitation
  learning and structured prediction,'' \emph{CoRR}, vol. abs/1011.0686, 2010.
  [Online]. Available: \url{http://arxiv.org/abs/1011.0686}
\BIBentrySTDinterwordspacing

\bibitem{9157239}
M.~Toromanoff, E.~Wirbel, and F.~Moutarde, ``End-to-end model-free
  reinforcement learning for urban driving using implicit affordances,'' in
  \emph{2020 IEEE/CVF Conference on Computer Vision and Pattern Recognition
  (CVPR)}.\hskip 1em plus 0.5em minus 0.4em\relax Los Alamitos, CA, USA: IEEE
  Computer Society, jun 2020, pp. 7151--7160.

\bibitem{chen2021learning}
D.~Chen, V.~Koltun, and P.~Kr{\"a}henb{\"u}hl, ``Learning to drive from a world
  on rails,'' in \emph{ICCV}, 2021.

\bibitem{rana2021bayesian}
K.~Rana, V.~Dasagi, J.~Haviland, B.~Talbot, M.~Milford, and N.~S{\"u}nderhauf,
  ``Bayesian controller fusion: Leveraging control priors in deep reinforcement
  learning for robotics,'' \emph{arXiv preprint arXiv:2107.09822}, 2021.

\bibitem{Daw2005UncertaintybasedCB}
N.~D. Daw, Y.~Niv, and P.~Dayan, ``Uncertainty-based competition between
  prefrontal and dorsolateral striatal systems for behavioral control,''
  \emph{Nature Neuroscience}, vol.~8, pp. 1704--1711, 2005.

\bibitem{Dayan2008DecisionTR}
P.~Dayan and N.~D. Daw, ``Decision theory, reinforcement learning, and the
  brain,'' \emph{Cognitive, Affective, \& Behavioral Neuroscience}, vol.~8, pp.
  429--453, 2008.

\bibitem{chen2019lbc}
D.~Chen, B.~Zhou, V.~Koltun, and P.~Kr\"ahenb\"uhl, ``Learning by cheating,''
  in \emph{Conference on Robot Learning (CoRL)}, 2019.

\bibitem{chitta2021neat}
K.~Chitta, A.~Prakash, and A.~Geiger, ``Neat: Neural attention fields for
  end-to-end autonomous driving,'' in \emph{International Conference on
  Computer Vision (ICCV)}, 2021.

\bibitem{isprs-annals-V-1-2021-145-2021}
\BIBentryALTinterwordspacing
Y.~Han and A.~Yilmaz, ``Dynamic routing for navigation in changing unknown maps
  using deep reinforcement learning,'' \emph{ISPRS Annals of the
  Photogrammetry, Remote Sensing and Spatial Information Sciences}, vol.
  V-1-2021, pp. 145--150, 2021. [Online]. Available:
  \url{https://www.isprs-ann-photogramm-remote-sens-spatial-inf-sci.net/V-1-2021/145/2021/}
\BIBentrySTDinterwordspacing

\bibitem{mnih2016asynchronous}
V.~Mnih, A.~P. Badia, M.~Mirza, A.~Graves, T.~Lillicrap, T.~Harley, D.~Silver,
  and K.~Kavukcuoglu, ``Asynchronous methods for deep reinforcement learning,''
  in \emph{Proceedings of The 33rd International Conference on Machine
  Learning}, ser. Proceedings of Machine Learning Research, M.~F. Balcan and
  K.~Q. Weinberger, Eds., vol.~48.\hskip 1em plus 0.5em minus 0.4em\relax New
  York, New York, USA: PMLR, 20--22 Jun 2016, pp. 1928--1937.

\bibitem{zhang2021endtoend}
Z.~Zhang, A.~Liniger, D.~Dai, F.~Yu, and L.~Van~Gool, ``End-to-end urban
  driving by imitating a reinforcement learning coach,'' in \emph{Proceedings
  of the IEEE/CVF International Conference on Computer Vision (ICCV)}, 2021.

\bibitem{inbook}
X.~Liang, T.~Wang, L.~Yang, and E.~Xing, \emph{CIRL: Controllable Imitative
  Reinforcement Learning for Vision-Based Self-driving: 15th European
  Conference, Munich, Germany, September 8–14, 2018, Proceedings, Part VII},
  09 2018, pp. 604--620.

\bibitem{dosovitskiy2017carla}
A.~Dosovitskiy, G.~Ros, F.~Codevilla, A.~Lopez, and V.~Koltun, ``{CARLA}: {An}
  open urban driving simulator,'' in \emph{Proceedings of the 1st Annual
  Conference on Robot Learning}, 2017, pp. 1--16.

\bibitem{he2015deep}
K.~He, X.~Zhang, S.~Ren, and J.~Sun, ``Deep residual learning for image
  recognition,'' in \emph{2016 IEEE Conference on Computer Vision and Pattern
  Recognition (CVPR)}, 2016, pp. 770--778.

\bibitem{haarnoja2018soft}
T.~Haarnoja, A.~Zhou, P.~Abbeel, and S.~Levine, ``Soft actor-critic: Off-policy
  maximum entropy deep reinforcement learning with a stochastic actor,'' in
  \emph{ICML}, 2018.

\bibitem{johannink2018residual}
T.~Johannink, S.~Bahl, A.~Nair, J.~Luo, A.~Kumar, M.~Loskyll, J.~A. Ojea,
  E.~Solowjow, and S.~Levine, ``Residual reinforcement learning for robot
  control,'' in \emph{2019 International Conference on Robotics and Automation
  (ICRA)}, 2019, pp. 6023--6029.

\bibitem{pertsch2020accelerating}
K.~Pertsch, Y.~Lee, and J.~J. Lim, ``Accelerating reinforcement learning with
  learned skill priors,'' in \emph{Conference on Robot Learning (CoRL)}, 2020.

\bibitem{sauer2018conditional}
A.~Sauer, N.~Savinov, and A.~Geiger, ``Conditional affordance learning for
  driving in urban environments,'' in \emph{Conference on Robot Learning
  (CoRL)}, 2018.

\bibitem{zhao2020sam}
A.~Zhao, T.~He, Y.~Liang, H.~Huang, G.~V. den Broeck, and S.~Soatto, ``Sam:
  Squeeze-and-mimic networks for conditional visual driving policy learning,''
  in \emph{Conference on Robot Learning (CoRL)}, 2020.

\end{thebibliography}
%



\end{document}